\pdfoutput=1  % 告知 arXiv 使用 pdflatex 编译
\documentclass{article}

\usepackage{microtype}
\usepackage{graphicx}
\usepackage{subcaption}
\usepackage{booktabs} 
\usepackage{hyperref}

\usepackage[preprint]{icml2026}

\usepackage{amsmath}
\usepackage{amssymb}
\usepackage{mathtools}
\usepackage{amsthm}

\usepackage[capitalize,noabbrev]{cleveref}

\theoremstyle{plain}

\theoremstyle{definition}

\theoremstyle{remark}

\usepackage[textsize=tiny]{todonotes}

\icmltitlerunning{See, Act, Adapt: Active Perception for Unsupervised Cross-Domain Visual Adaptation}

\usepackage{listings}
\usepackage{xcolor}
\definecolor{bg}{rgb}{0.96,0.96,0.96}
\lstdefinestyle{wideprompt}{
    backgroundcolor=\color{bg},
    basicstyle=\ttfamily\small,
    breaklines=true,
    breakatwhitespace=true,
    keywordstyle=\color{blue},
    stringstyle=\color{teal},
    commentstyle=\color{gray},
    frame=single,
    framerule=0pt,
    framesep=10pt,
    showstringspaces=false,
    captionpos=b,
}
\definecolor{GainGreen}{RGB}{34,139,34}
\definecolor{LossRed}{RGB}{178,34,34}
\newcommand{\gain}[1]{\textcolor{GainGreen}{\small\,(#1)}}
\newcommand{\loss}[1]{\textcolor{LossRed}{\small\,(#1)}}
\usepackage{siunitx}
\usepackage{makecell}

\begin{document}

\twocolumn[
  \icmltitle{See, Act, Adapt: Active Perception for Unsupervised Cross-Domain Visual Adaptation via Personalized VLM-Guided Agent}

  % --- 作者列表 ---
  \begin{icmlauthorlist}
    \icmlauthor{Tianci Tang}{zju}
    \icmlauthor{Tielong Cai}{zju}
    \icmlauthor{Hongwei Wang}{zju}
    \icmlauthor{Gaoang Wang}{zju}
  \end{icmlauthorlist}

  % --- 单位信息 ---
  \icmlaffiliation{zju}{Zhejiang University, Hangzhou, China}

  % --- 联系方式（根据需要修改邮箱） ---
  \icmlcorrespondingauthor{Tianci Tang}{tianci.24@intl.zju.edu.cn}
  \icmlcorrespondingauthor{Gaoang Wang}{gaoangwang@intl.zju.edu.cn}

  \icmlkeywords{Machine Learning, Active Perception, Cross-Domain Adaptation, VLM}

  \vskip 0.3in
]

% 显示单位脚注
\printAffiliationsAndNotice{} 

% --- 正文内容 ---
\begin{abstract}
Pre-trained perception models excel in generic image domains but degrade significantly in novel environments like indoor scenes. The conventional remedy is fine-tuning on downstream data which incurs catastrophic forgetting of prior knowledge and demands costly, scene-specific annotations. We propose a paradigm shift through Sea$^2$ (\textbf{Se}e, \textbf{A}ct, \textbf{A}dapt): rather than adapting the perception modules themselves, we adapt how they are deployed through an intelligent pose-control agent. Sea$^2$ keeps all perception modules frozen, requiring no downstream labels during training, and uses only scalar perceptual feedback to navigate the agent toward informative viewpoints. 
Specially, we transform a vision-language model (VLM) into a low-level pose controller through a two-stage training pipeline: first fine-tuning it on rule-based exploration trajectories that systematically probe indoor scenes, and then refining the policy via unsupervised reinforcement learning that constructs rewards from the perception module’s outputs and confidence.
Unlike prior active perception methods that couple exploration with specific models or collect data for retraining them, Sea$^2$ directly leverages off-the-shelf perception models for various tasks without the need for retraining.
We conducted experiments on three visual perception tasks, including visual grounding, segmentation and 3D box estimation, with performance improvements of 13.54\%, 15.92\% and 27.68\% respectively on dataset ReplicaCAD.
\end{abstract}    
\section{Introduction}

\begin{figure}[t]
  \centering
  \includegraphics[width=\linewidth]{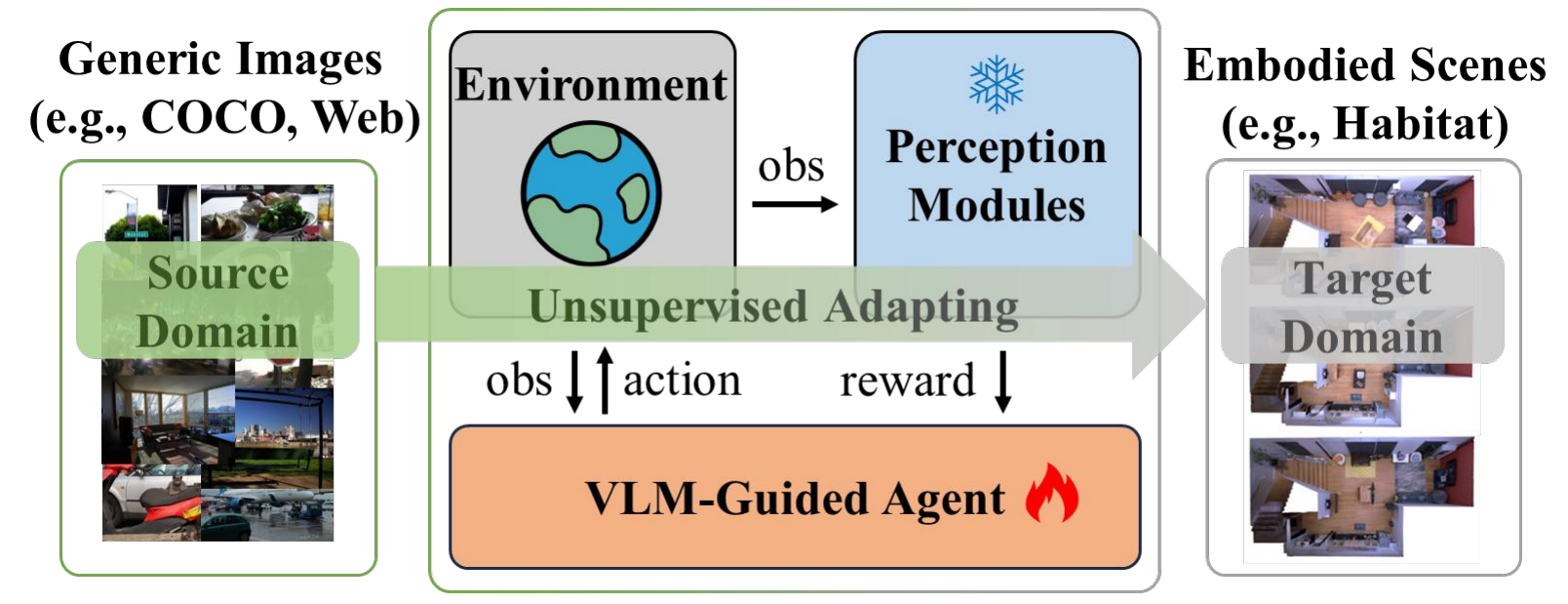}
  \caption{\textbf{Overview of unsupervised cross-domain adaptation via VLM-guided active perception agent.} (Left) Source domain including generic images (e.g., COCO~\cite{lin2015microsoftcococommonobjects}, Web) where perception models pre-trained on. (Middle) Our embodied agent actively explores the indoor environment and adjusts its camera pose to capture information-rich observations that maximize perception quality. (Right) Target domain including embodied scenes (e.g., Habitat~\cite{savva2019habitat}) where our VLM-guided agent trained on to unsupervised adapt the domain gap.}
  \label{fig:overview}
\end{figure}

\begin{figure*}[t]
    \centering
    \includegraphics[width=0.95\textwidth]{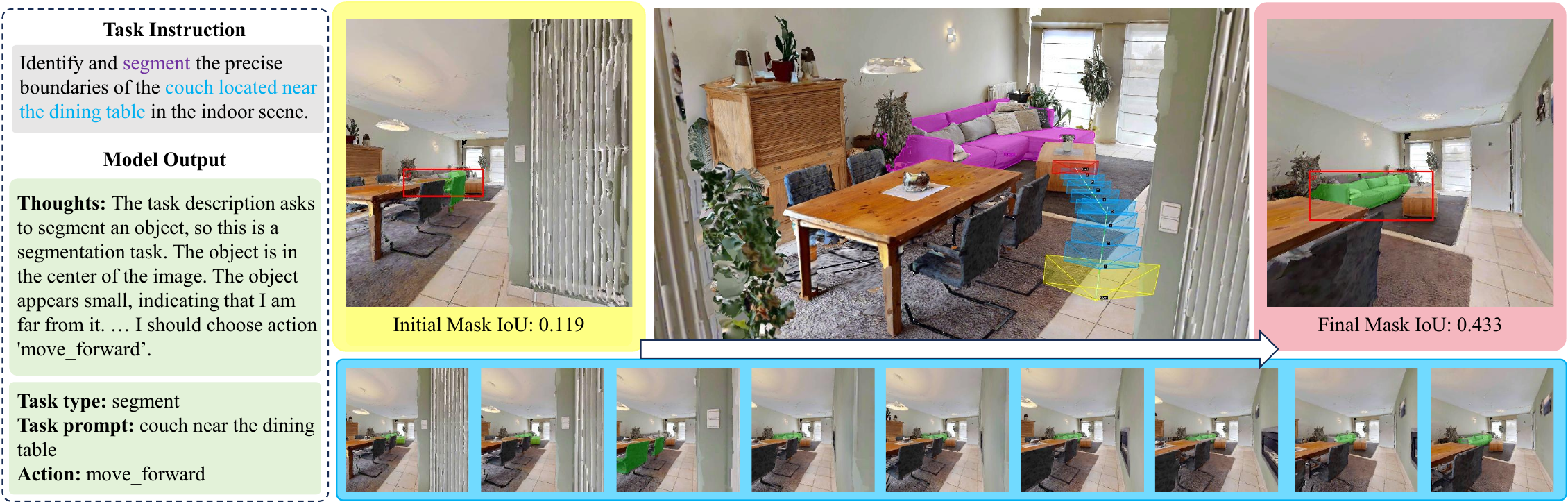}
    \caption{\textbf{Illustration of the active perception process for segmentation.}
      The agent decomposes the instruction into task-specific metadata and reasons about the scene to execute camera-adjusting actions. Starting from a highly occluded initial view (yellow) where the perception output (green mask) for the target object (red box) is poor, it follows a trajectory of navigational steps (blue) to reduce visual ambiguity. The final viewpoint (red) offers a significantly clearer perspective for the perception module, yielding a substantial improvement in perception score compared to the initial state.}
    \label{fig:demo}
\end{figure*}

Large-scale visual models pre-trained on internet-scale imagery have demonstrated remarkable generalization across recognition, segmentation, and visual grounding tasks~\cite{xiao2024florence,xiao2023clip,ravi2024sam,ren2024groundedsamassemblingopenworld,He2021MaskedAA,10657649}. However, when deployed in novel embodied environments such as indoor scenes, their performance degrades sharply due to domain gaps in viewpoint distribution, occlusion patterns, and spatial semantics~\cite{yang2019embodied,ammirato2017dataset,zeng2020view,fan2024evidential}. The prevailing remedy is fine-tuning perception modules on downstream data, while it incurs two critical limitations: catastrophic forgetting of prior knowledge and the prohibitive cost of acquiring scene-specific annotations (e.g., pixel masks, 3D bounding boxes, or referring expressions). This raises a fundamental question: \textit{Can we adapt perception to new domains without touching the models themselves?}

We propose a paradigm shift: instead of adapting perception modules themselves, we adapt how they are deployed. Our key insight is that perceptual performance depends not only on model capacity but also critically on the informativeness of the observation viewpoint~\cite{yang2019embodied,fan2024evidential}. By intelligently controlling the agent’s pose to seek out informative views, we can recover performance drops without retraining, annotating data, or modifying the perception system. Crucially, our approach freezes all perception modules, requiring no downstream labels during agent training; it relies solely on scalar feedback derived directly from the frozen perception module (e.g., detection confidence, segmentation quality, or 3D consistency). This enables operation in open-world settings where ground-truth annotations are unavailable, unlike prior active perception methods that either tightly couple exploration with specific architectures~\cite{chaplot2021seal,kotar2022interactron,ding2023learning} or collect labeled data for retraining~\cite{scarpellini2024look,jing2023learning}. Moreover, unlike recent RL-based approaches that assume closed-set tasks or rely on task-specific uncertainty models~\cite{ding2023learning,fan2024evidential}, our framework leverages semantic reasoning to handle open-ended, natural-language-driven objectives (e.g., “Locate the bicycle next to the door.”) while generalizing across diverse perception backbones.

To realize this vision, we transform a vision-language model (VLM) into a low-level pose controller through a two-stage training pipeline. First, we align the VLM with spatial reasoning via supervised fine-tuning on rule-based exploration trajectories that systematically probe indoor environments. Second, we refine the policy using unsupervised reinforcement learning (RL), where rewards are constructed from the outputs of the frozen perception module without access to any downstream perceptual annotations (e.g., ground-truth masks or bounding boxes). This decouples perception from control, creating a modular framework that seamlessly accommodates diverse off-the-shelf perception architectures without the need for task-specific retraining.

We validate our approach on three mainstream vision tasks on datasets ReplicaCAD~\cite{szot2021habitat} and HM3D~\cite{ramakrishnan2021hm3d} in photo-realistic Habitat~\cite{savva2019habitat} including: visual grounding, segmentation, and 3D bounding box estimation. Sea$^2$ improves performance on ReplicaCAD and HM3D by 13.54\%, 15.92\%, and 27.68\% and 22.16\%, 12.49\%, and 9.31\%, respectively, demonstrating that strategic viewpoint selection can recover domain-gap-induced degradation without a single annotated label. This establishes a new direction for label-efficient domain adaptation in embodied AI.

Our contributions can be summarized as follows:

\begin{itemize}
% \item We present the first VLM-based active perception framework that achieves zero-shot transfer across perception architectures and scenes by freezing all models and using only their scalar outputs as reward signals, eliminating the need for downstream annotations or model updates.
\item We present the first VLM-based active perception framework that achieves plug-and-play compatibility with diverse off-the-shelf models. By using only scalar outputs as rewards, Sea$^2$ enables seamless integration across various perception architectures without requiring retraining or downstream labels.

\item We introduce an unsupervised RL training pipeline based on perception-derived rewards. By leveraging only task-level objectives and scalar outputs from frozen models, our method eliminates the need for dense perceptual annotations (e.g., pixel-level masks or 3D boxes), enabling effective policy learning in annotation-scarce environments.

\item We demonstrate substantial gains across three visual tasks, including detection, segmentation, and 3D understanding, with improvements of 13.54\%, 15.92\%, and 27.68\% in metrics on dataset ReplicaCAD, showing that viewpoint adaptation alone can effectively bridge domain gaps.
\end{itemize}
\section{Related Work}
\label{sec:related work}

\subsection{Embodied Active Perception}
Active perception studies how agents move to acquire views that improve recognition under occlusion and partial visibility. Early \emph{Embodied Visual Recognition (EVR)} work showed that learned motion policies can outperform passive baselines for classification, amodal localization, and segmentation in 3D environments~\cite{yang2019embodied}. Subsequent lines either adapt the detector in-the-loop during interaction~\cite{kotar2022interactron}, or keep perception frozen and optimize the \emph{policy} to seek informative views, e.g., decision-transformer control with offline+online training~\cite{ding2023learning}. Complementary efforts collect/evaluate informative multi-view data for self-training or exploration-aware adaptation~\cite{fang2020move,chaplot2021seal,jing2023learning,scarpellini2024look}, incorporate uncertainty to guide viewpoint selection~\cite{fan2024evidential}, extend from detection to active localization~\cite{di2024learning}, and address embodied domain adaptation under distribution shift~\cite{shi2025embodied}. 
Our work (i) keeps all perception modules \emph{frozen} to avoid catastrophic forgetting, (ii) uses a single VLM policy that conditions on both images and natural-language task prompts to \emph{control viewpoints} across multiple modules.

\subsection{Multimodal Foundation Models}
Multimodal foundation models provide open-vocabulary recognition and promptable perception. CLIP~\cite{radford2021learning} aligns images and text for robust zero-shot recognition; BLIP-2~\cite{li2023blip} bridges frozen vision encoders and LLMs for efficient multimodal reasoning; Flamingo scales few-shot VLMs; instruction-tuned MLLMs such as LLaVA~\cite{liu2023visual} improve visual instruction following. Recent families (Qwen3-VL~\cite{yang2025qwen3}, InternVL3.5~\cite{wang2025internvl3}) further enhance grounding and OCR across parameter scales. Recent work has utilized the multimodal perception and planning capabilities of VLM for embodied navigation~\cite{zhang2024navid,zhang2024uni,jiang2025alphadrive,ao2025emacembodiedmultimodalagent,yin2024sgnavonline3dscene,2025arXiv250213451Z} rather than purely for perception or generation. Inspired by this, we \emph{use a VLM as an action policy} to translate language-grounded spatial reasoning into low-level control while keeping task modules frozen.

\subsection{RL-based Post-training for VLMs}
RLHF learns a reward model from human preferences, then optimize the policy with PPO under a KL constraint (InstructGPT)~\cite{ouyang2022training}; strong alignment but costly and reward-sensitive. DPO removes reward modeling and directly fit a preference objective from pairs~\cite{rafailov2023direct}; simpler but still preference-dependent. GRPO-style objectives have recently improved long-horizon reasoning and training stability~\cite{shao2024deepseekmath,zheng2025group,yu2025dapo}. We adopt GRPO to train the VLM as an action policy from \emph{unsupervised} rewards avoiding preference data and detector fine-tuning while enabling transferable, language-guided active perception.

\begin{figure*}[t]
  \centering
  \includegraphics[width=0.85\linewidth]{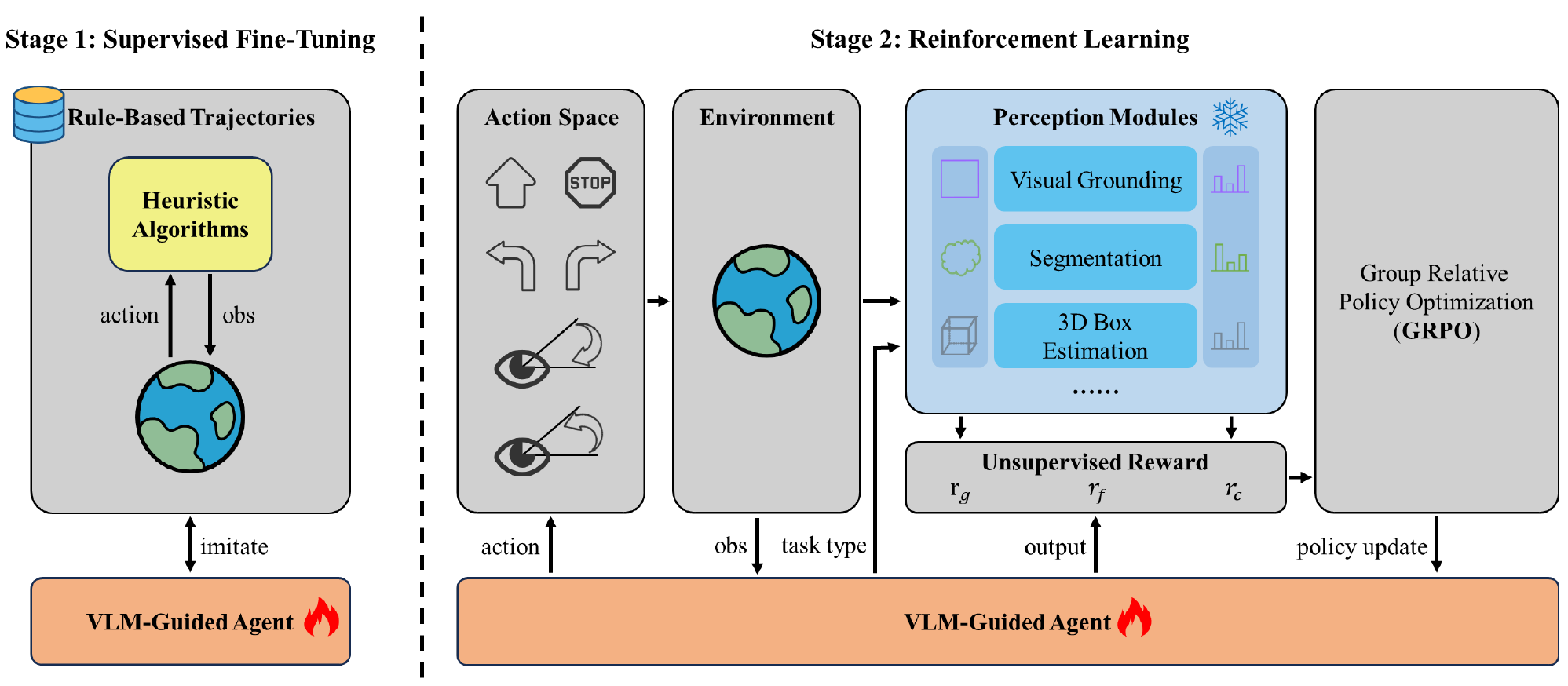}
  \caption{\textbf{Illustration of our Sea$^2$ framework.} In Stage 1, the VLM is fine-tuned on rule-based trajectories generated by heuristic algorithms to align it with spatial reasoning and control formats. In Stage 2, the VLM serves as a low-level pose controller for the agent, where it is further refined using unsupervised reinforcement learning with GRPO.  The agent interacts with the environment, receiving observations and taking actions to optimize its policy based on rewards derived from the frozen selected perception module's confidence and results (e.g., grounding confidence, mask area).  The selected perception module remain frozen throughout the training process, ensuring no catastrophic forgetting of prior knowledge.  The final policy enables the agent to navigate to informative viewpoints that enhance the performance of the perception modules without requiring any downstream annotations.}
  \label{fig:framework}
\end{figure*}

\begin{figure*}[t]
    \centering
    % Example segmentation figure – replace the filename below
    \includegraphics[width=0.95\textwidth]{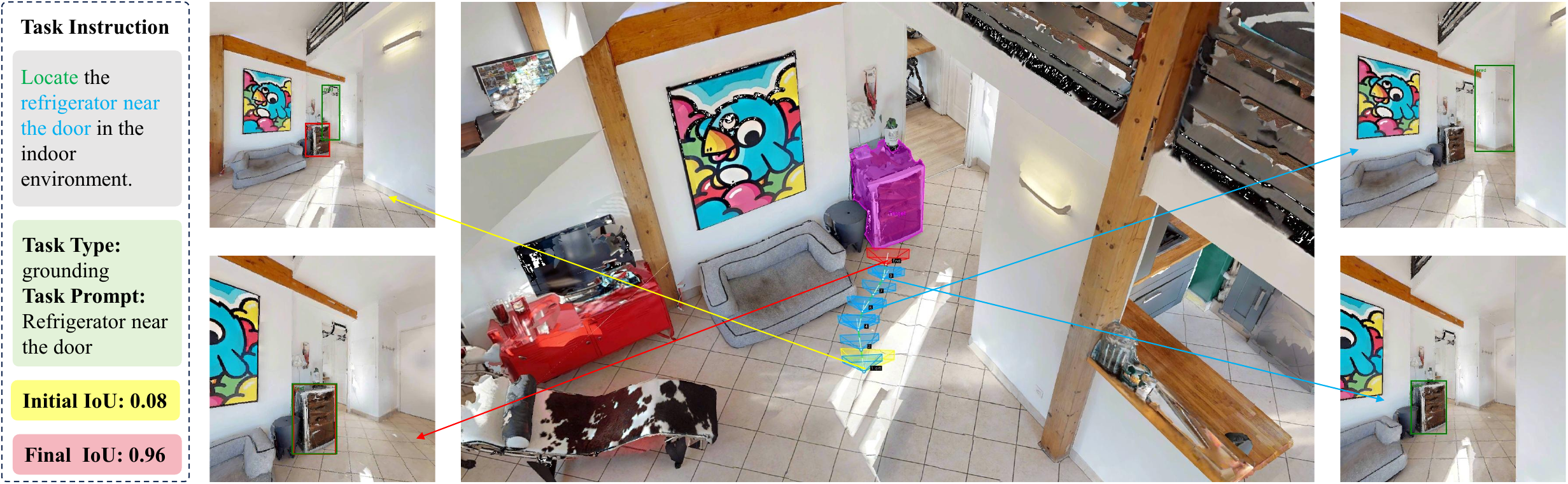}

    \caption{
        \textbf{Illustration of the active perception process for visual grounding.} From a poor initial view (yellow) where the prediction (green box) for the target (red box) is inaccurate, the agent takes navigational steps (blue) to reduce ambiguity, reaching a final viewpoint (red) that greatly improves the perception result.
    }
    \label{fig:seg_sofa_example}
\end{figure*}

\begin{figure*}[t]
    \centering
    \includegraphics[width=0.95\textwidth]{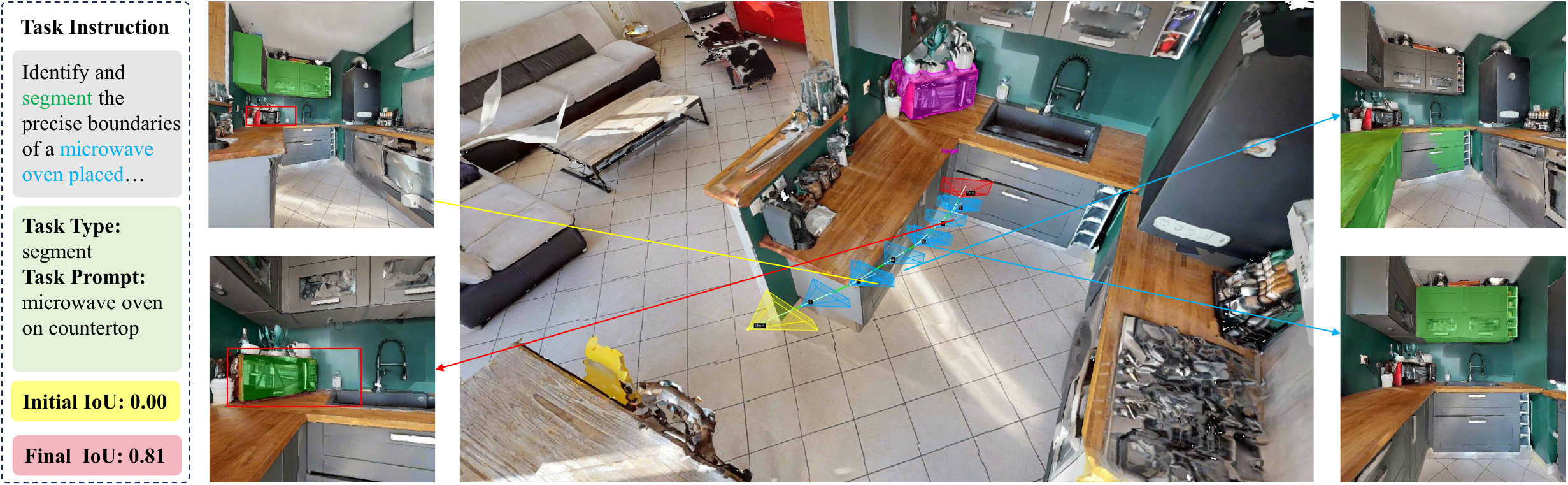}

    \caption{
        \textbf{Illustration of the active perception process for segmentation.} From a poor initial view (yellow) where the prediction (green box) for the target (red box) is inaccurate, the agent takes navigational steps (blue) to reduce ambiguity, reaching a final viewpoint (red) that greatly improves the perception result.
    }
    % \label{fig:seg_sofa_example}
\end{figure*}

\begin{figure*}[t]
    \centering
    % Example segmentation figure – replace the filename below
    \includegraphics[width=0.95\textwidth]{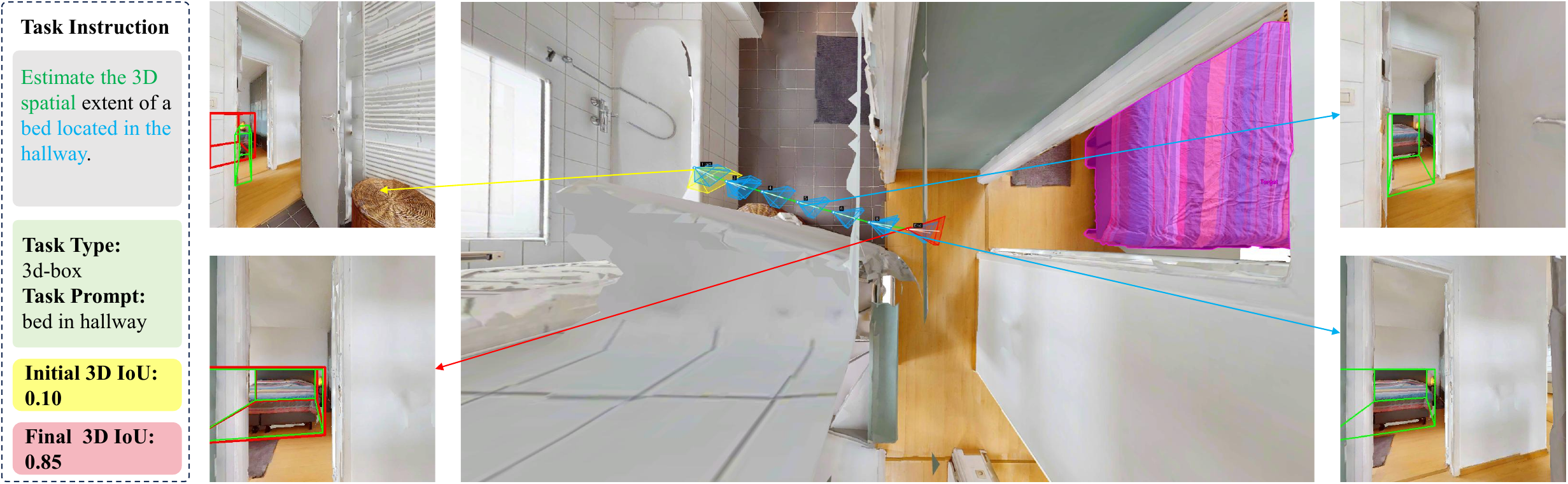}

    \caption{
        \textbf{Illustration of the active perception process for 3D box estimation.} From a poor initial view (yellow) where the prediction (green box) for the target (red box) is inaccurate, the agent takes navigational steps (blue) to reduce ambiguity, reaching a final viewpoint (red) that greatly improves the perception result.
    }
    % \label{fig:seg_sofa_example}
\end{figure*}
% =========== ReplicaCAD ===========
\begin{table*}[t]
\centering
\small
\setlength{\tabcolsep}{2.5pt}
\caption{Three perception module baselines on ReplicaCAD with different controllers. Visual grounding is evaluated by mAP, segmentation by IoU and Dice, and 3D box estimation by IoU and Center Score (higher is better). The best results are highlighted in \textbf{bold}. Degraded results are highlighted in red, while improved results are highlighted in green. \textbf{Notably, the Shortest Path baseline is a ground-truth-informed (oracle) method: it possesses perfect prior knowledge of target locations, retrieving precise 3D coordinates from scene annotations and computing the shortest navigable path using the Habitat pathfinder.}}
\label{tab:train-result-replicaCAD}
\begin{tabular}{l
S[table-format=1.4]
S[table-format=1.4]
S[table-format=1.4]
S[table-format=1.4]
S[table-format=1.4]
S[table-format=1.4]
S[table-format=1.4]
}
\toprule
& \multicolumn{3}{c}{\textbf{Visual Grounding}}
& \multicolumn{2}{c}{\textbf{Segmentation}}
& \multicolumn{2}{c}{\textbf{3D Box Estimation}}\\
\cmidrule(lr){2-4}\cmidrule(lr){5-6}\cmidrule(lr){7-8}
\textbf{Perception Module + Policy}
& \multicolumn{1}{c}{mAP@0.5 $\uparrow$}
& \multicolumn{1}{c}{mAP@0.75 $\uparrow$}
& \multicolumn{1}{c}{mAP$_\text{avg}$ $\uparrow$}
& \multicolumn{1}{c}{IoU $\uparrow$}
& \multicolumn{1}{c}{Dice $\uparrow$}
& \multicolumn{1}{c}{IoU $\uparrow$}
& \multicolumn{1}{c}{Center Score $\uparrow$}
\\
\midrule
\textcolor{gray}{Pretrained Perception Module (PPM)}
& {\textcolor{gray}{\makecell{\num{0.7958}}}}
& {\textcolor{gray}{\makecell{\num{0.6225}}}}
& {\textcolor{gray}{\makecell{\num{0.7092}}}}
& {\textcolor{gray}{\makecell{\num{0.5621}}}}
& {\textcolor{gray}{\makecell{\num{0.6398}}}}
& {\textcolor{gray}{\makecell{\num{0.2648}}}}
& {\textcolor{gray}{\makecell{\num{0.5499}}}}
\\
PPM + Forward
& {\makecell{\num{0.4667}\\ \loss{-41.37\%}}}
& {\makecell{\num{0.3897}\\ \loss{-37.40\%}}}
& {\makecell{\num{0.4282}\\ \loss{-39.62\%}}}
& {\makecell{\num{0.3495}\\ \loss{-37.82\%}}}
& {\makecell{\num{0.3911}\\ \loss{-38.87\%}}}
& {\makecell{\num{0.1503}\\ \loss{-43.24\%}}}
& {\makecell{\num{0.3206}\\ \loss{-41.70\%}}}
\\
PPM + Random
& {\makecell{\num{0.6662}\\ \loss{-16.29\%}}}
& {\makecell{\num{0.5183}\\ \loss{-16.74\%}}}
& {\makecell{\num{0.5923}\\ \loss{-16.48\%}}}
& {\makecell{\num{0.4626}\\ \loss{-17.70\%}}}
& {\makecell{\num{0.5282}\\ \loss{-17.44\%}}}
& {\makecell{\num{0.2237}\\ \loss{-15.48\%}}}
& {\makecell{\num{0.4809}\\ \loss{-12.55\%}}}
\\
PPM + Heuristic
& {\makecell{\num{0.6969}\\ \loss{-12.43\%}}}
& {\makecell{\num{0.5698}\\ \loss{-8.47\%}}}
& {\makecell{\num{0.6334}\\ \loss{-10.69\%}}}
& {\makecell{\num{0.5200}\\ \loss{-7.49\%}}}
& {\makecell{\num{0.5849}\\ \loss{-8.58\%}}}
& {\makecell{\num{0.2543}\\ \loss{-3.93\%}}}
& {\makecell{\num{0.5310}\\ \loss{-3.44\%}}}
\\
PPM + Shortest Path
& {\makecell{\num{0.8668}\\ \gain{+8.92\%}}}
& {\makecell{\num{0.7198}\\ \gain{+15.64\%}}}
& {\makecell{\num{0.7933}\\ \gain{+11.86\%}}}
& {\makecell{\num{0.6075}\\ \gain{+8.07\%}}}
& {\makecell{\num{0.6817}\\ \gain{+6.55\%}}}
& {\makecell{\num{0.3259}\\ \gain{+23.09\%}}}
& {\makecell{\num{0.6373}\\ \gain{+15.89\%}}}
\\
PPM + InternVL3.5-2B
& {\makecell{\num{0.7020}\\ \loss{-11.79\%}}}
& {\makecell{\num{0.5747}\\ \loss{-7.68\%}}}
& {\makecell{\num{0.6384}\\ \loss{-9.99\%}}}
& {\makecell{\num{0.5556}\\ \loss{-1.16\%}}}
& {\makecell{\num{0.6322}\\ \loss{-1.19\%}}}
& {\makecell{\num{0.2528}\\ \loss{-4.53\%}}}
& {\makecell{\num{0.5250}\\ \loss{-4.52\%}}}
\\
PPM + Qwen3VL-2B
& {\makecell{\num{0.6133}\\ \loss{-22.93\%}}}
& {\makecell{\num{0.5097}\\ \loss{-18.13\%}}}
& {\makecell{\num{0.5615}\\ \loss{-20.83\%}}}
& {\makecell{\num{0.4214}\\ \loss{-25.03\%}}}
& {\makecell{\num{0.4744}\\ \loss{-25.85\%}}}
& {\makecell{\num{0.1781}\\ \loss{-32.74\%}}}
& {\makecell{\num{0.4038}\\ \loss{-26.57\%}}}
\\
\textbf{PPM + Ours (InternVL3.5-2B)}
& {\makecell{\num{0.8627}\\ \gain{+8.40\%}}}
& {\makecell{\num{0.7055}\\ \gain{+13.33\%}}}
& {\makecell{\num{0.7841}\\ \gain{+10.56\%}}}
& {\makecell{\num{0.6251}\\ \gain{+11.21\%}}}
& {\makecell{\num{0.7002}\\ \gain{+9.44\%}}}
& {\makecell{\num{0.3195}\\ \gain{+20.67\%}}}
& {\makecell{\num{0.6600}\\ \gain{+20.03\%}}}
\\
\textbf{PPM + Ours (Qwen3VL-2B)}
& {\makecell{\textbf{0.8725}\\ \gain{+9.63\%}}}
& {\makecell{\textbf{0.7378}\\ \gain{+18.53\%}}}
& {\makecell{\textbf{0.8052}\\ \gain{+13.54\%}}}
& {\makecell{\textbf{0.6516}\\ \gain{+15.92\%}}}
& {\makecell{\textbf{0.7267}\\ \gain{+13.59\%}}}
& {\makecell{\textbf{0.3380}\\ \gain{+27.68\%}}}
& {\makecell{\textbf{0.6893}\\ \gain{+25.35\%}}}
\\
\bottomrule
\end{tabular}
\vspace{-0.5em}
\end{table*}

% ======== HM3D =========
\begin{table*}[t]
\centering
\small
\setlength{\tabcolsep}{2.5pt}
\caption{Three perception module baselines on HM3D with different controllers. Visual grounding is evaluated by mAP, segmentation by IoU and Dice, and 3D box estimation by IoU and Center Score (higher is better). The best results are highlighted in \textbf{bold}. Degraded results are highlighted in red, while improved results are highlighted in green.}
\label{tab:train-result-hm3d}
\begin{tabular}{l
S[table-format=1.4]
S[table-format=1.4]
S[table-format=1.4]
S[table-format=1.4]
S[table-format=1.4]
S[table-format=1.4]
S[table-format=1.4]
}
\toprule
& \multicolumn{3}{c}{\textbf{Visual Grounding}}
& \multicolumn{2}{c}{\textbf{Segmentation}}
& \multicolumn{2}{c}{\textbf{3D Box Estimation}}\\
\cmidrule(lr){2-4}\cmidrule(lr){5-6}\cmidrule(lr){7-8}
\textbf{Perception Module + Policy}
& \multicolumn{1}{c}{mAP@0.5 $\uparrow$}
& \multicolumn{1}{c}{mAP@0.75 $\uparrow$}
& \multicolumn{1}{c}{mAP$_\text{avg}$ $\uparrow$}
& \multicolumn{1}{c}{IoU $\uparrow$}
& \multicolumn{1}{c}{Dice $\uparrow$}
& \multicolumn{1}{c}{IoU $\uparrow$}
& \multicolumn{1}{c}{Center Score $\uparrow$}
\\
\midrule
\textcolor{gray}{Pretrained Perception Module (PPM)}
& {\textcolor{gray}{\makecell{\num{0.5914}}}}
& {\textcolor{gray}{\makecell{\num{0.4224}}}}
& {\textcolor{gray}{\makecell{\num{0.5069}}}}
& {\textcolor{gray}{\makecell{\num{0.4694}}}}
& {\textcolor{gray}{\makecell{\num{0.5460}}}}
& {\textcolor{gray}{\makecell{\num{0.3136}}}}
& {\textcolor{gray}{\makecell{\num{0.5653}}}}
\\
PPM + Forward
& {\makecell{\num{0.3133}\\ \loss{-47.02\%}}}
& {\makecell{\num{0.2475}\\ \loss{-41.40\%}}}
& {\makecell{\num{0.2804}\\ \loss{-44.68\%}}}
& {\makecell{\num{0.2362}\\ \loss{-49.69\%}}}
& {\makecell{\num{0.2700}\\ \loss{-50.55\%}}}
& {\makecell{\num{0.1429}\\ \loss{-54.42\%}}}
& {\makecell{\num{0.2879}\\ \loss{-49.08\%}}}
\\
PPM + Random
& {\makecell{\num{0.5032}\\ \loss{-14.91\%}}}
& {\makecell{\num{0.3739}\\ \loss{-11.48\%}}}
& {\makecell{\num{0.4386}\\ \loss{-13.48\%}}}
& {\makecell{\num{0.4052}\\ \loss{-13.68\%}}}
& {\makecell{\num{0.4689}\\ \loss{-14.12\%}}}
& {\makecell{\num{0.2714}\\ \loss{-13.45\%}}}
& {\makecell{\num{0.4866}\\ \loss{-13.92\%}}}
\\
PPM + Heuristic
& {\makecell{\num{0.5403}\\ \loss{-8.64\%}}}
& {\makecell{\num{0.4043}\\ \loss{-4.28\%}}}
& {\makecell{\num{0.4723}\\ \loss{-6.83\%}}}
& {\makecell{\num{0.4754}\\ \gain{+1.26\%}}}
& {\makecell{\num{0.5384}\\ \loss{-1.39\%}}}
& {\makecell{\num{0.3343}\\ \gain{+6.60\%}}}
& {\makecell{\num{0.5700}\\ \gain{+0.83\%}}}
\\
PPM + Shortest Path
& {\makecell{\num{0.5997}\\ \gain{+1.40\%}}}
& {\makecell{\num{0.4988}\\ \gain{+18.09\%}}}
& {\makecell{\num{0.5493}\\ \gain{+8.35\%}}}
& {\makecell{\num{0.4915}\\ \gain{+4.71\%}}}
& {\makecell{\num{0.5614}\\ \gain{+2.82\%}}}
& {\makecell{\num{0.3316}\\ \gain{+5.74\%}}}
& {\makecell{\num{0.5853}\\ \gain{+3.53\%}}}
\\
PPM + InternVL3.5-2B
& {\makecell{\num{0.4512}\\ \loss{-23.71\%}}}
& {\makecell{\num{0.3606}\\ \loss{-14.62\%}}}
& {\makecell{\num{0.4059}\\ \loss{-19.93\%}}}
& {\makecell{\num{0.4745}\\ \gain{+1.09\%}}}
& {\makecell{\num{0.5526}\\ \gain{+1.21\%}}}
& {\makecell{\num{0.3028}\\ \loss{-3.43\%}}}
& {\makecell{\num{0.5482}\\ \loss{-3.03\%}}}
\\
PPM + Qwen3VL-2B
& {\makecell{\num{0.4512}\\ \loss{-23.71\%}}}
& {\makecell{\num{0.3606}\\ \loss{-14.62\%}}}
& {\makecell{\num{0.4059}\\ \loss{-19.93\%}}}
& {\makecell{\num{0.4326}\\ \loss{-7.85\%}}}
& {\makecell{\num{0.5007}\\ \loss{-8.29\%}}}
& {\makecell{\num{0.2761}\\ \loss{-11.95\%}}}
& {\makecell{\num{0.4725}\\ \loss{-16.42\%}}}
\\
\textbf{PPM + Ours (InternVL3.5-2B)}
& {\makecell{\num{0.6400}\\ \gain{+8.21\%}}}
& {\makecell{\num{0.5230}\\ \gain{+23.82\%}}}
& {\makecell{\num{0.5815}\\ \gain{+14.72\%}}}
& {\makecell{\num{0.4981}\\ \gain{+6.12\%}}}
& {\makecell{\num{0.5624}\\ \gain{+3.01\%}}}
& {\makecell{\num{0.3306}\\ \gain{+5.42\%}}}
& {\makecell{\num{0.5748}\\ \gain{+1.67\%}}}
\\
\textbf{PPM + Ours (Qwen3VL-2B)}
& {\makecell{\textbf{0.6752}\\ \gain{+14.17\%}}}
& {\makecell{\textbf{0.5633}\\ \gain{+33.37\%}}}
& {\makecell{\textbf{0.6193}\\ \gain{+22.16\%}}}
& {\makecell{\textbf{0.5261}\\ \gain{+12.49\%}}}
& {\makecell{\textbf{0.5908}\\ \gain{+8.57\%}}}
& {\makecell{\textbf{0.3420}\\ \gain{+9.31\%}}}
& {\makecell{\textbf{0.5955}\\ \gain{+5.67\%}}}
\\
\bottomrule
\end{tabular}
\vspace{-0.5em}
\end{table*}

\section{Method}
\label{sec:method}
Our core premise is that perceptual performance degradation in novel domains stems not only from model capacity limitations, but also from suboptimal viewpoint selection~\cite{yang2019embodied,fan2024evidential}. Rather than adapting perception modules, which risks catastrophic forgetting and demands costly annotations, we freeze all perception models and instead learn an unsupervised pose-control policy that navigates toward informative observations. The policy is optimized solely through scalar feedback from the frozen modules (confidence scores and geometric consistency), requiring no downstream labels during training. In this section, we introduce our framework design, which is an indoor VLM-guided embodied agent with three perception modules. We also elaborate on a two-stage training pipeline for our proposed method.

\subsection{Overview and Problem Formulation}
We formulate the task as an unsupervised active perception problem where an embodied agent controls its camera pose to maximize the quality of observations for a set of frozen perception modules $\mathcal{M}$. At episode onset, the agent receives a natural-language task instruction $I$ and an initial observation $o_1$. The agent's policy $\pi_\theta$ generates the task prompt $p$, selects the perception module $m \in \mathcal{M}$ and outputs discrete actions $a_t \in \mathcal{A}$ over a horizon $T$ to adjust its viewpoint, where $\mathcal{A}$ includes translational and rotational movements (e.g., move forward, turn right/left, look up/down, stop).

The observation $o_t$ at time $t$ is an RGB image captured from the current pose. The frozen selected perception module $m$ processes $(o_t, p)$ to produce a prediction $\hat{y}_t^m$ and a scalar confidence score $c_t^m \in [0, 1]$ indicating result reliability. Critically, no ground-truth labels are available for $\hat{y}_t^m$; the agent must infer viewpoint quality solely from the sequence of $(\hat{y}_t^m, c_t^m)$.

\paragraph{Optimization Objective.} The agent's goal is to learn a policy $\pi_\theta$ that maximizes the cumulative quality of perception outcomes without updating any module parameters:
\begin{equation}
    \max_{\theta} \mathbb{E}_{\pi_\theta} \left[ \sum_{t=1}^{T} r\big(\{\hat{y}_t^m, c_t^m\}_{m \in \mathcal{M}}, o_t\big) \right]
\end{equation}

where the reward $r(\cdot)$ is an unsupervised scalar signal derived entirely from the frozen modules' outputs and geometric consistency checks. This decouples perception from control, enabling zero-shot transfer across module architectures and scenes.

\subsection{Framework Design}

Sea$^2$ employs a VLM as the action policy $\pi_\theta$. The VLM receives a task instruction $I$ (encoding the original task type $h_I$ and language description $p_I$ of the target object, further details regarding the formulation of $I$ are provided in the \textbf{Task Definition} (see \ref{para:task_def}).) and the current observation $o_t$, then generates a structured output comprising:
\begin{itemize}
\item Thoughts: A textual reasoning trace explaining the spatial reasoning (e.g., object location, occlusion assessment).
\item Task type: A task routing result $h$ used to select perception module $m \in \mathcal{M}$.
\item Task prompt: A language description $p$ of the task for the selected perception module $m$.
\item Action: A discrete control command $a_t \in \mathcal{A}$.
\end{itemize}
The VLM is transformed from a passive reasoning model into an embodied pose controller through our two-stage training pipeline, \textit{i.e.}, supervised fine-tuning (SFT) followed by reinforcement learning (RL).

\paragraph{Perception Module Interface.} Each module $m \in \mathcal{M}$ implements a unified interface: given $(o_t, p)$, it returns a prediction $\hat{y}_t^m$ (\textit{e.g.}, grounded 2D box, segmentation mask, or 3D box) and a confidence score $c_t^m$. The confidence is based on each perception module. During both SFT and RL, all modules remain frozen, treating them as black-box experts that provide feedback without adaptation.

\subsection{Training Pipeline}
As shown in \cref{fig:framework}, the first stage involves supervised fine-tuning (SFT) of the VLM using rule-based trajectories. The second stage employs reinforcement learning (RL) with reward signals derived from perception modules' feedback.

\subsubsection{Supervised Fine-Tuning (SFT)} 
To initialize the VLM with basic spatial reasoning and reduce RL exploration variance, we collect trajectories using a deterministic heuristic policy. The heuristic follows a three-phase logic:
\begin{itemize}
\item Object Search: Rotate until the target object is detected with non-zero confidence.
\item Viewpoint Centering: Adjust the viewpoint to align the object's predicted region with the image center.
\item Proximity Adjustment: Move forward until the object occupies a sufficient image area, then stop.
\end{itemize}
Each trajectory records the heuristic's ``thoughts" (spatial reasoning), task type, task prompt and actions, forming a supervised dataset for SFT that aligns the VLM's output format with embodied control requirements.

\subsubsection{Reinforcement Learning (RL)} 
After acquiring fundamental embodied perception skills through SFT, 
we further improve the model via reinforcement learning (RL) using perception-based reward signals. 
In this stage, we adopt the Group Relative Policy Optimization (GRPO) algorithm and design a comprehensive reward function to guide policy optimization. 
The total reward is composed of a \textit{format reward} $r_{f}$, a \textit{confidence reward} $r_{c}$, 
and a \textit{geometric reward} $r_{g}$. 
Among them, $r_{f}$ is derived from the action policy output, 
while $r_{c}$ and $r_{g}$ are computed from the perception module outputs. 
The overall reward function is defined as:

\begin{equation}
  r = \begin{cases} 
    r_{f} + \lambda_{1}r_{c} + \lambda_{2}r_{g}, & \text{if } h = h_I \\
    -1, & \text{if } h \neq h_I
  \end{cases}
  \label{eq:reward}
\end{equation}
where $h$ and $h_I$ denote the predicted and ground-truth task types, respectively. The agent receives a constant penalty of $-1$ for incorrect task identification; otherwise, the reward is defined as a weighted combination of the fundamental perception score $r_f$, the confidence score $r_c$, and the geometric consistency $r_g$. $\lambda_1$ and $\lambda_2$ weight the respective rewards and satisfy $\lambda_1+\lambda_2=1$.
Specifically, while the agent utilizes the ground-truth task type $h_I$ to determine the reward structure, the training process remains entirely independent of perceptual ground-truth annotations. This ensures that the policy learns to optimize observation quality using only the intrinsic feedback of the frozen perception models, making it applicable to open-world scenarios where precise spatial or semantic labels are unavailable.

\paragraph{Format Reward.}
The format reward $r_{f}$ evaluates whether the model’s output structure adheres to the expected schema. 
Specifically, the reasoning process must be contained in the \texttt{"thoughts":\{\}} field, 
the action decision must appear in the \texttt{"action":\{\}} field, 
and only one action can be generated per step. 
This reward encourages the model to output well-structured and consistent reasoning–action pairs:
\begin{equation}
r_{f} =
\begin{cases}
0.05, & \text{if the format is correct},\\
-0.05, & \text{if the format is incorrect}.
\end{cases}
\end{equation}

\paragraph{Confidence Reward.}
The confidence reward $r_{c}$ measures the change in the perception module’s confidence score between consecutive steps, 
encouraging the agent to increase confidence in its perception results without relying on external supervision. 
For visual grounding, the confidence is obtained from GroundingDINO; 
for segmentation, it is a weighted combination of GroundingDINO and SAM; 
and for 3D box estimation, it integrates confidence from GroundingDINO, SAM, and the 3D box estimator.
\begin{equation}
  r_{c} = c_{t}^{m} - c_{t-1}^{m}
  \label{eq:confidence reward}
\end{equation}

\paragraph{Geometric Reward.}
The geometric reward $r_{g}$ enforces spatial consistency between the predicted region and the observation. 
It consists of two components: \textit{area} and \textit{center}.

The area reward measures the proportion of the predicted region $\hat{y}_{t}^{m}$ within the current observation $o_t$, 
encouraging the model to approach the target object. 
For visual grounding, this ratio corresponds to the bounding-box area relative to the image; 
for segmentation, to the mask area; 
and for 3D box estimation, to the projected 3D box area.
\begin{equation}
  r_{a} = \frac{A(\hat{y}_{t}^{m})}{A(o_t)}
  \label{eq:area reward}
\end{equation}
where $A(\cdot)$ is the area function.

The center reward encourages the model to align the predicted target region with the image center. 
Instead of using raw Euclidean distance, we compute a normalized alignment score between 0 and 1, 
where higher values indicate better alignment. 
This design stabilizes localization behavior and prevents excessive deviation from the target region during training.
\begin{equation}
r_{u} = 1 - d\big(u(\hat{y}_t^{m}),\, u(o_t)\big)
\label{eq:center reward}
\end{equation}
where $u(\cdot)$ extracts the predicted and observed center coordinates and $d(\cdot,\, \cdot)$ computes normalized distance. 

The geometric reward $r_{g}$ calculation method is as follows:
\begin{align}
r_{g} &= g_{t}^{m} - g_{t-1}^{m} \\
g_{t}^{m} &= \mu_1 r_{a} + \mu_2 r_{u}
\end{align}
where $g_t^m$ is the geometric score and $\mu_1$ and $\mu_2$ are the weights of individual rewards.

\begin{table*}[t]
\centering
\small
\setlength{\tabcolsep}{2.5pt}
\caption{Ablation study on training strategies. Visual grounding is evaluated by mAP, segmentation by IoU and Dice, and 3D box estimation by IoU and Center Score (higher is better). The best results are highlighted in \textbf{bold}. Degraded results are highlighted in red, while improved results are highlighted in green.}
\label{tab:ablation-train-new}
\begin{tabular}{l
S[table-format=1.4]
S[table-format=1.4]
S[table-format=1.4]
S[table-format=1.4]
S[table-format=1.4]
S[table-format=1.4]
S[table-format=1.4]
}
\toprule
& \multicolumn{3}{c}{\textbf{Visual Grounding}}
& \multicolumn{2}{c}{\textbf{Segmentation}}
& \multicolumn{2}{c}{\textbf{3D Box Estimation}}\\
\cmidrule(lr){2-4}\cmidrule(lr){5-6}\cmidrule(lr){7-8}
\textbf{Method}
& \multicolumn{1}{c}{mAP@0.5 $\uparrow$}
& \multicolumn{1}{c}{mAP@0.75 $\uparrow$}
& \multicolumn{1}{c}{mAP$_\text{avg}$ $\uparrow$}
& \multicolumn{1}{c}{IoU $\uparrow$}
& \multicolumn{1}{c}{Dice $\uparrow$}
& \multicolumn{1}{c}{IoU $\uparrow$}
& \multicolumn{1}{c}{Center Score $\uparrow$}
\\
\midrule
\textcolor{gray}{Pretrained Perception Module}
& {\textcolor{gray}{\makecell{\num{0.7958}}}}
& {\textcolor{gray}{\makecell{\num{0.6225}}}}
& {\textcolor{gray}{\makecell{\num{0.7092}}}}
& {\textcolor{gray}{\makecell{\num{0.5621}}}}
& {\textcolor{gray}{\makecell{\num{0.6398}}}}
& {\textcolor{gray}{\makecell{\num{0.2648}}}}
& {\textcolor{gray}{\makecell{\num{0.5499}}}}
\\
RL-Only
& {\makecell{\num{0.4647}\\ \loss{-41.61\%}}}
& {\makecell{\num{0.3877}\\ \loss{-37.72\%}}}
& {\makecell{\num{0.4262}\\ \loss{-39.90\%}}}
& {\makecell{\num{0.3513}\\ \loss{-37.50\%}}}
& {\makecell{\num{0.3921}\\ \loss{-38.71\%}}}
& {\makecell{\num{0.1511}\\ \loss{-42.92\%}}}
& {\makecell{\num{0.3239}\\ \loss{-41.09\%}}}
\\
SFT-Only
& {\makecell{\num{0.8027}\\ \gain{+0.86\%}}}
& {\makecell{\num{0.6738}\\ \gain{+8.25\%}}}
& {\makecell{\num{0.7383}\\ \gain{+4.10\%}}}
& {\makecell{\num{0.5997}\\ \gain{+6.70\%}}}
& {\makecell{\num{0.6719}\\ \gain{+5.02\%}}}
& {\makecell{\num{0.3099}\\ \gain{+17.05\%}}}
& {\makecell{\num{0.6012}\\ \gain{+9.32\%}}}
\\
\textbf{SFT+RL}
& {\makecell{\textbf{0.8627}\\ \gain{+8.40\%}}}
& {\makecell{\textbf{0.7055}\\ \gain{+13.33\%}}}
& {\makecell{\textbf{0.7841}\\ \gain{+10.56\%}}}
& {\makecell{\textbf{0.6251}\\ \gain{+11.21\%}}}
& {\makecell{\textbf{0.7002}\\ \gain{+9.44\%}}}
& {\makecell{\textbf{0.3195}\\ \gain{+20.67\%}}}
& {\makecell{\textbf{0.6600}\\ \gain{+20.03\%}}}
\\
\bottomrule
\end{tabular}
\vspace{-0.5em}
\end{table*}

\begin{table*}[t]
\centering
\small
\setlength{\tabcolsep}{2.5pt}
\caption{Ablation study on reward functions. Visual grounding is evaluated by mAP, segmentation by IoU and Dice, and 3D box estimation by IoU and Center Score (higher is better). Degraded results are highlighted in red, while improved results are highlighted in green.} 
\label{tab:ablation-reward-new}
\begin{tabular}{l
S[table-format=1.4]
S[table-format=1.4]
S[table-format=1.4]
S[table-format=1.4]
S[table-format=1.4]
S[table-format=1.4]
S[table-format=1.4]
}
\toprule
& \multicolumn{3}{c}{\textbf{Visual Grounding}}
& \multicolumn{2}{c}{\textbf{Segmentation}}
& \multicolumn{2}{c}{\textbf{3D Box Estimation}}\\
\cmidrule(lr){2-4}\cmidrule(lr){5-6}\cmidrule(lr){7-8}
\textbf{Reward function}
& \multicolumn{1}{c}{mAP@0.5 $\uparrow$}
& \multicolumn{1}{c}{mAP@0.75 $\uparrow$}
& \multicolumn{1}{c}{mAP$_\text{avg}$ $\uparrow$}
& \multicolumn{1}{c}{IoU $\uparrow$}
& \multicolumn{1}{c}{Dice $\uparrow$}
& \multicolumn{1}{c}{IoU $\uparrow$}
& \multicolumn{1}{c}{Center Score $\uparrow$}
\\
\midrule
\textcolor{gray}{Format}
& {\textcolor{gray}{\makecell{---}}}
& {\textcolor{gray}{\makecell{---}}}
& {\textcolor{gray}{\makecell{---}}}
& {\textcolor{gray}{\makecell{---}}}
& {\textcolor{gray}{\makecell{---}}}
& {\textcolor{gray}{\makecell{---}}}
& {\textcolor{gray}{\makecell{---}}}
\\
Format + Geometric
& {\makecell{\num{0.8292}\\ \gain{+4.20\%}}}
& {\makecell{\num{0.6872}\\ \gain{+10.39\%}}}
& {\makecell{\num{0.7582}\\ \gain{+6.91\%}}}
& {\makecell{\num{0.5942}\\ \gain{+5.71\%}}}
& {\makecell{\num{0.6667}\\ \gain{+4.20\%}}}
& {\makecell{\num{0.2905}\\ \gain{+9.71\%}}}
& {\makecell{\num{0.6125}\\ \gain{+11.38\%}}}
\\
Format + Confidence
& {\makecell{\num{0.7115}\\ \loss{-10.60\%}}}
& {\makecell{\num{0.5562}\\ \loss{-10.65\%}}}
& {\makecell{\num{0.6339}\\ \loss{-10.62\%}}}
& {\makecell{\num{0.5253}\\ \loss{-6.54\%}}}
& {\makecell{\num{0.5973}\\ \loss{-6.64\%}}}
& {\makecell{\num{0.2902}\\ \gain{+9.59\%}}}
& {\makecell{\num{0.5954}\\ \gain{+8.27\%}}}
\\
Format + Confidence + Geometric
& {\makecell{\textbf{\num{0.8627}}\\ \gain{+8.40\%}}}
& {\makecell{\textbf{\num{0.7055}}\\ \gain{+13.33\%}}}
& {\makecell{\textbf{\num{0.7841}}\\ \gain{+10.56\%}}}
& {\makecell{\textbf{\num{0.6251}}\\ \gain{+11.21\%}}}
& {\makecell{\textbf{\num{0.7002}}\\ \gain{+9.44\%}}}
& {\makecell{\textbf{\num{0.3195}}\\ \gain{+20.67\%}}}
& {\makecell{\textbf{\num{0.6600}}\\ \gain{+20.03\%}}}
\\
\bottomrule
\end{tabular}
\vspace{-0.5em}
\end{table*}

\section{Experiments}
\label{sec:experiments}

\subsection{Experimental Setup and Data}
\label{sec:exp-setup}

\paragraph{Task Definition.}
\label{para:task_def}
At the beginning of each episode, the agent is spawned at a random navigable pose with a forward-facing RGB camera (height $1$\,m, $512{\times}512$ resolution). We select as target an object that is (partially) in view and exceeds a minimum visible-area threshold to avoid degenerate language references. The task instruction $I$ is synthesized via a two-stage hierarchical process. First, we employ \texttt{Qwen2.5-VL-7B}~\cite{bai2025qwen2} to generate an open-vocabulary, attribute-aware language description $p_I$ of the target object. Subsequently, the final task instruction $I$ is formulated by conditioning on the ground-truth task type $h_I$ (e.g., ``segment the [$p_I$]''), thereby encoding both the target's semantic attributes and the specific perceptual objective. The policy is allowed up to $T=10$ steps and may issue \texttt{stop} early once it believes the current viewpoint is informative. Formally, given observation $o_t$ and prompt $p$, the VLM-based policy $\pi$ chooses an action $a_t{=}\pi(o_t,p)$ and we record the perception output sequence $(\hat{y}^m, c^m)$ for each frozen module $m\in\mathcal{M}$.

\paragraph{Environments and Splits.}
All experiments are conducted in \texttt{Habitat}~\cite{savva2019habitat} on \texttt{ReplicaCAD}~\cite{szot2021habitat} suite (84 indoor scenes) and \texttt{HM3D}~\cite{ramakrishnan2021hm3d} suite. For each suite, we randomly select 10 scenes for training and sample 500 tasks from the 10 training scenes to construct the test sets for evaluation. 

\paragraph{Action Space and Observations.}
We use a low-level discrete action set 
with step sizes of $0.25$\,m (translation) and $10^\circ$ (yaw/pitch). Vertical look controls are included to handle tall/near objects and reduce self-occlusion. Each step provides the current RGB frame; depth is used only by the 3D module (below), not as an input channel to the policy.

\paragraph{Perception Modules.}
We consider three representative perception tasks and their modules $\mathcal{M}{=}\{\texttt{Visual Grounding}, \texttt{Segmentation},\\ \texttt{3D Box Estimation}\}$; each returns a prediction and a scalar confidence $conf_m$ used for feedback and rewards.
\emph{(i) Visual Grounding} (VG) uses \texttt{GroundingDINO}~\cite{liu2024groundingdinomarryingdino} (GDINO) to produce a 2D bounding box and confidence $(\hat{y}^{vg},\,c^{vg})$ given the language description and the current RGB frame.
\emph{(ii) Segmentation} first applies \texttt{GroundingDINO} to obtain a box and then \texttt{SAM}~\cite{kirillov2023segment} to produce a mask $(\hat{y}^{seg})$. Its confidence is a weighted combination of detector and mask confidences $c^{seg}{=}\mu_{3}c^{seg1}{+}\mu_{4}c^{seg2}$ (fixed weights across experiments).
\emph{(iii) 3D Box Estimation} computes a point cloud by intersecting the mask with the depth image and fits an oriented 3D bounding box via minimum-area footprint search on the ground plane. Its confidence aggregates detector, mask, and geometric signals (e.g., point count, filtering retention, depth consistency) into $c^{3D}\in[0,1]$. 
% All three modules are used as-is (no fine-tuning) and their confidences are not calibrated post hoc.

\paragraph{SFT Data Collection.}
To align the pretrained VLM with embodied control and reduce RL exploration burden, we collect supervised trajectories with a deterministic heuristic that is agnostic to perception architectures:
\emph{Search}---rotate until the target becomes visible;
\emph{Centering}---partition the image into a $3{\times}3$ grid; if the target center is off-axis, issue $\{\texttt{turn\_left}, \texttt{turn\_right}, \texttt{look\_up}, \texttt{look\_down}\}$ to re-center;
\emph{Approach/Stop}---once centered, if the visible region is below a threshold, \texttt{move\_forward}; otherwise \texttt{stop}.
At each step we log the textual \texttt{"thoughts"} and the \texttt{"action"} token. These trajectories seed the \emph{SFT} stage.

\paragraph{GRPO Training Details.}
For reinforcement learning, we optimize the policy with GRPO using the following hyperparameters:
initial learning rate $1{\times}10^{-6}$, $4$ rollouts per update, batch size $8$, KL coefficient $0.04$, and sampling temperature $0.4$.
Rewards are computed from frozen-module feedback as defined in \cref{eq:reward} (confidence change in \cref{eq:confidence reward} and geometric shaping in \cref{eq:center reward}).

\paragraph{Baselines and Metrics.}

We compare against: \emph{(a) Pretrained Perception Module (PPM)}---evaluate each frozen module on the first frame at $t{=}1$ (no action); \emph{(b) Forward}---always execute \texttt{move\_forward}; \emph{(c) Random}---sample a random action from $\mathcal{A}$ at each step; \emph{(d) Heuristic}---the search--centering--approach heuristic used for SFT data collection; \emph{(e) Shortest Path}---a baseline that has access to ground-truth object positions, serving as a reference for evaluating learned policies under the same constraints. 

For baselines (a)--(e), the ground-truth task type $h_I$ and target description $p_I$ are directly provided as inputs. In contrast, for \emph{(f) InternVL3.5-2B~\cite{wang2025internvl3} \& Qwen3VL-2B~\cite{yang2025qwen3}} (prompt-based VLM controllers) and \emph{(g) Ours (InternVL3.5) \& Ours (Qwen3VL)} (trained via our two-stage pipeline), the models receive only the natural language instruction $I$. These models must autonomously parse the task type $h$ and object description $p$ from $I$. To ensure rigorous evaluation, if a model misidentifies the task type ($h \neq h_I$), the corresponding metric for that episode is set to zero; furthermore, any distortion in the extracted $p$ naturally impacts the downstream perception module's performance.

We use AP for visual grounding, IoU/Dice for segmentation, and IoU/center score for 3D box estimation. Metrics are computed from the final module outputs at \texttt{stop} (or $t{=}T$). Baselines and metrics correspond to \cref{tab:train-result-replicaCAD}.

%%%%%%%%%%%%%%%%%%%%%%%%%%%%%%%%%%%%%%%%%%%  实验图表

\begin{figure}[t]
  \centering
  \includegraphics[width=1.0\columnwidth]{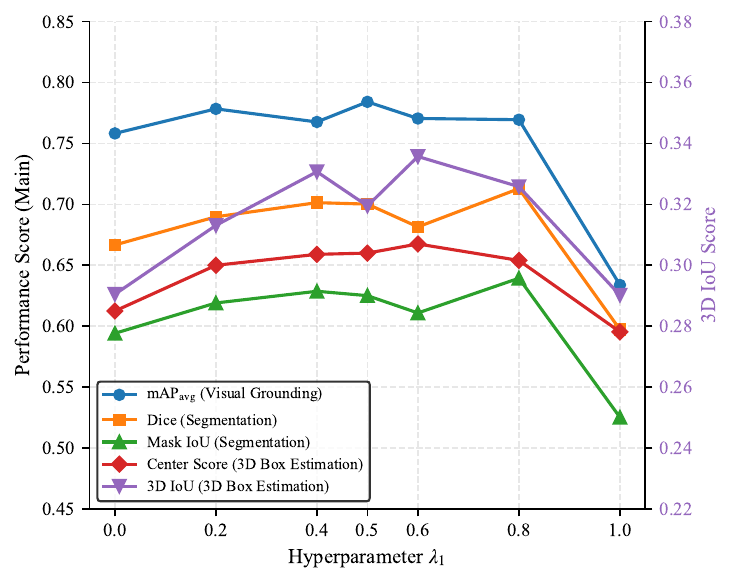}
  \caption{\small Impact of hyperparameter $\lambda_1$ on different tasks. The left y-axis corresponds to Grounding and Segmentation metrics, while the right y-axis highlights the 3D IoU performance.}
  \label{fig:lambda_ablation}
\end{figure}

\subsection{Performance in Perception Tasks}
\label{sec:perf}

Following the setup and baselines defined in \cref{sec:exp-setup}, \cref{tab:train-result-replicaCAD} summarizes results on ReplicaCAD across visual grounding, segmentation, and 3D box estimation with frozen modules. Relative to the \texttt{PPM} baseline, our policy boosts VG AP$_{avg}$ {$+13.54\%$}, segmentation IoU {$+15.92\%$} with Dice rising {$+13.59\%$}, and 3D box IoU {$+27.68\%$}, 3D center score {$+25.35\%$}.

Simple motion baselines are ineffective or even harmful. \texttt{Forward} policy substantially degrades all metrics (e.g., $-41.37\%$ seg IoU), likely due to over-approaching that increases occlusion or truncation. \texttt{Random} motion is also inferior across tasks, underscoring that \emph{which} view to acquire matters. 

The \texttt{Heuristic} baseline improves segmentation and 3D box metrics only marginally (e.g., $+3.37\%$ seg IoU) and lags significantly behind our policy. This is primarily because \texttt{Heuristic} is purely reactive, relying entirely on the immediate feedback of the frozen perception module; consequently, any initial misdetection becomes unrecoverable, leading the agent into erroneous trajectories. Meanwhile, although \texttt{Shortest path} possesses privileged knowledge of the target's 3D coordinates to ensure geometric reachability, its performance gains remain modest (e.g., $+6.27\%$ 3D IoU). This indicates that simply arriving at the target's location is insufficient for high-quality perception. Unlike \texttt{Shortest path}, which lacks the capacity for viewpoint-level planning, Sea$^2$ strategically optimizes the camera pose to mitigate occlusion and maximize visual informativeness.

Finally, directly prompting a compact VLM controller 

without task-aligned training underperforms the static initial value on all metrics, highlighting that naively using a VLM as a policy is insufficient; embodiment alignment and reward-driven refinement are necessary. 

We further evaluate Sea$^2$ on HM3D, a dataset featuring more complex and high-fidelity 3D-scanned reconstructions of real-world environments. As summarized in \cref{tab:train-result-hm3d}, the performance gains on HM3D mirror those observed on ReplicaCAD, reinforcing the effectiveness and robustness of our active perception strategy across diverse scene distributions.

\subsection{Ablation Study}
\label{sec:ablation}

\paragraph{Training Strategy.}
\Cref{tab:ablation-train-new} compares \emph{RL-Only}, \emph{SFT-Only}, and our \emph{SFT+RL}. Training the policy from scratch with RL is unstable and underperforms the initial value on recognition metrics. SFT on rule-based trajectories yields consistent gains across tasks, validating the importance of a spatially grounded cold start. The full \emph{SFT+RL} pipeline achieves the best results on all metrics, showing that SFT provides stable priors while RL unlocks task-specific view selection beyond heuristic behaviors.

\paragraph{Reward Design.}
\Cref{tab:ablation-reward-new} further disentangles the contribution of each reward term in \cref{eq:reward}. Using \emph{confidence} alone (\emph{Format+Confidence}) is unstable: it degrades visual grounding and segmentation, while offering improvements only on 3D box estimation. Although confidence is a strong {directional} signal toward better viewpoints, it is also noisy and volatile; the effect is most pronounced in VG, where the signal comes solely from \texttt{GroundingDINO}, making the policy more susceptible to detector-specific jitter.
By contrast, the \emph{geometric} term (\emph{Format+Geometric}) is {stable} and {module-agnostic}. Optimizing area/center alignment (\cref{eq:center reward}) yields consistent but limited improvements across tasks, reflecting that geometric shaping provides reliable spatial guidance yet lacks model/task specificity.
Crucially, {combining} the two signals (\emph{Format+Confidence+Geometric}) produces {large} and {uniform} gains. Geometry supplies smooth, low-variance spatial shaping, while confidence calibrates view selection to the current frozen module and task, turning a noisy-but-informative signal into a robust control objective. 
\cref{fig:lambda_ablation} analyzes the impact of $\lambda_1$, indicating a necessary trade-off between the reward components. Consequently, we fix $\lambda_1 = 0.5$ for all main evaluations.

\section{Conclusions}
\label{sec:conclusions}
We presented our Sea$^2$ that adapts pre-trained perception modules to new embodied environments without updating parameters or requiring downstream labels. By learning a pose-control policy through supervised heuristic trajectories and unsupervised GRPO-based RL, the agent actively selects informative viewpoints using only scalar feedback from frozen perception modules. Experiments on visual grounding, segmentation, and 3D box estimation demonstrate gains, 13.54\%, 15.92\%, and 27.68\% respectively on ReplicaCAD, highlighting that intelligently controlling viewpoints offers an efficient and annotation-free alternative to traditional model fine-tuning for domain adaptation.

% \section*{Impact Statement}
% This work advances fundamental research in active perception and embodied AI by enabling frozen vision models to adapt to new environments through intelligent viewpoint selection without requiring labeled data or model updates...

% 参考文献
\bibliography{main}
\bibliographystyle{icml2026}

%%%%%%%%%%%%%%%%%%%%%%%%%%%%%%%%%%%%%%%%%%%%%%%%%%%%%%%%%%%%%%%%%%%%%%%%%%%%%%%
% 附录
%%%%%%%%%%%%%%%%%%%%%%%%%%%%%%%%%%%%%%%%%%%%%%%%%%%%%%%%%%%%%%%%%%%%%%%%%%%%%%%
% \newpage
% \appendix
% \onecolumn
% \input{sec/X_appendix}

\end{document}